\documentclass[]{article}
\usepackage{amsmath, amssymb, amsthm,bbm}
\usepackage{algorithm, algorithmic}
\usepackage{geometry}

\RequirePackage[colorlinks,citecolor=blue]{hyperref}

\RequirePackage[numbers]{natbib}

\usepackage{orcidlink}

\theoremstyle{plain}
\newtheorem{theorem}{Theorem}
\newtheorem{assume}{Assumption}

\newtheorem{corollary}{Corollary}

\title{Exponential Lasso: robust sparse penalization under heavy-tailed noise and outliers with exponential-type loss}
\author{The Tien Mai\orcidlink{0000-0002-3514-9636}}

\date{
\small
Norwegian Institute of Public Health, Oslo, 0456, Norway
\\
email: the.tien.mai@fhi.no
}

\begin{document}

\maketitle

\begin{abstract}
In high-dimensional statistics, the Lasso is a cornerstone method for simultaneous variable selection and parameter estimation. However, its reliance on the squared loss function renders it highly sensitive to outliers and heavy-tailed noise, potentially leading to unreliable model selection and biased estimates. To address this limitation, we introduce the Exponential Lasso, a novel robust method that integrates an exponential-type loss function within the Lasso framework. This loss function is designed to achieve a smooth trade-off between statistical efficiency under Gaussian noise and robustness against data contamination. Unlike other methods that cap the influence of large residuals, the exponential loss smoothly redescends, effectively downweighting the impact of extreme outliers while preserving near-quadratic behavior for small errors. We establish theoretical guarantees showing that the Exponential Lasso achieves strong statistical convergence rates, matching the classical Lasso under ideal conditions while maintaining its robustness in the presence of heavy-tailed contamination. Computationally, the estimator is optimized efficiently via a Majorization-Minimization (MM) algorithm that iteratively solves a series of weighted Lasso subproblems. Numerical experiments demonstrate that the proposed method is highly competitive, outperforming the classical Lasso in contaminated settings and maintaining strong performance even under Gaussian noise.
 
Our method is implemented in the \texttt{R} package \texttt{heavylasso} available on Github: \url{https://github.com/tienmt/heavylasso}.
\end{abstract}


\section{Introduction}

In modern data analysis, it is common to encounter datasets where the number of features ($p$) greatly exceeds the number of samples ($n$), a setting that breaks down classical statistical methods. When faced with this high-dimensionality, the traditional least squares estimator becomes unreliable and ill-posed. This has spurred the development of regularization methods, which are designed to impose structure, prevent overfitting, and improve the model's interpretability by favoring simpler, sparser solutions \citep{hastie2009elements,buhlmann2011statistics,giraud2021introduction}.

A pioneering and widely adopted technique in this domain is the Lasso (least absolute shrinkage and selection operator) \citep{tibshirani1996regression}. It provides a powerful framework for performing both variable selection and parameter estimation simultaneously. The Lasso finds the coefficient vector $\boldsymbol{\beta}$ by solving the following optimization problem:
$$
\widehat{\boldsymbol{\beta}}^{\text{lasso}}
=
\arg\min_{\boldsymbol{\beta} \in \mathbb{R}^p}
\left\{
\frac{1}{2n} \|\mathbf{y} - \mathbf{X}\boldsymbol{\beta} \|_2^2
+
\lambda \|\boldsymbol{\beta} \|_1
\right\}
$$
In this formulation, the first term is the standard least squares loss function, while the second is an $L_1$ penalty weighted by a tuning parameter $\lambda > 0$. By penalizing the sum of the absolute values of the coefficients, the Lasso effectively shrinks many of them to exactly zero, thus performing automated feature selection. The foundational principles of the Lasso have ignited a vast body of research, leading to numerous advancements in sparse estimation and high-dimensional inference \citep{zou2006adaptive,bunea2007sparsity,zhao2006model,lounici2008sup,bellec2018slope}.

However, the squared loss underlying the classical Lasso implicitly assumes light-tailed, approximately Gaussian errors. In many real-world datasets—such as those arising in genomics, finance, and environmental monitoring—this assumption is frequently violated. Outliers or heavy-tailed noise can exert a disproportionate influence on the squared loss, 
leading to biased estimates and poor variable selection performance \citep{loh2017statistical,loh2024theoretical}. To mitigate such sensitivity, numerous robust variants of the Lasso have been proposed, often by replacing the squared loss with more robust alternatives such as 
the Huber loss \citep{sardy2001robust,loh2017statistical,loh2021scale}, 
Tukey’s biweight loss \citep{chang2018robust,smucler2017robust}, 
Student's loss \citep{mai2025heavy},
or rank-based and median-of-means losses \citep{rejchel2020rank,lecue2020robust,wang2020tuning}. These methods reduce the impact of extreme residuals but may introduce additional tuning complexity or require nontrivial optimization schemes when the loss becomes nonconvex.

In this work, we consider a novel robust loss function—the exponential-type loss—within the Lasso penalization framework, designed to achieve a smooth trade-off between efficiency under Gaussian noise and robustness under heavy-tailed contamination. The proposed estimator, termed the Exponential Lasso, is defined as
\[
\widehat{\boldsymbol{\beta}}
=
\arg\min_{\boldsymbol{\beta} \in \mathbb{R}^p}
\left\{
\frac{1}{n}
\sum_{i=1}^n
\frac{1}{\tau}\left[
1 - \exp \left(
-\frac{\tau (y_i - \mathbf{x}_i^\top \boldsymbol{\beta})^2}{2}
\right)\right]
+
\lambda \|\boldsymbol{\beta} \|_1
\right\},
  \]
  where \(\tau > 0\) controls the degree of robustness. When \(\tau \to 0\), the exponential loss approaches the squared loss, recovering the classical Lasso. For finite \(\tau\), large residuals are exponentially downweighted, effectively limiting their influence on parameter estimation.

The intuition behind this exponential-type loss is simple yet powerful: it penalizes small residuals nearly quadratically—preserving statistical efficiency under light-tailed noise—while suppressing the contribution of extreme deviations through the exponential term. In contrast to the Huber loss, which caps the linear growth of large residuals, the exponential loss smoothly redescends, assigning progressively smaller weights to extreme outliers. This property aligns with the influence-function perspective in robust statistics \citep{hampel1974influence}, where bounded and redescending functions provide strong resistance to contamination. As a result, the Exponential Lasso achieves both robustness and differentiability, enabling efficient optimization and stability in high-dimensional regimes.

From a computational standpoint, the exponential loss admits a natural Majorization–Minimization (MM)
algorithmic interpretation. Each iteration reweights the residuals according to their exponential downweighting factor, leading to a sequence of weighted Lasso subproblems. This results in a fast, stable, and interpretable algorithm with minimal modification to standard Lasso solvers.

Our theoretical results establishe that the proposed Exponential Lasso estimator achieves reliable estimation accuracy even in high-dimensional settings and under the presence of outliers or heavy-tailed noise. 
The theoretical result guarantees that, with an appropriate choice of the tuning parameter, the estimator attains the same convergence rate as the classical Lasso under ideal conditions. This robustness is ensured under a mild assumption on the noise distribution, requiring only that the errors have a positive probability of lying within a central region rather than having light tails. The proof combines a local curvature argument showing that the exponential loss remains well-behaved near the true parameter with concentration techniques that control random fluctuations in the data \citep{loh2017statistical,loh2021scale}. 

To demonstrate the effectiveness of our method, we conduct extensive simulation studies under a variety of settings involving heavy-tailed noise and outliers. We compare our approach to several Lasso variants that employ different loss functions, 
including the squared loss \citep{tibshirani1996regression}, $\ell_1$ loss \citep{wang2007robust}, Huber loss \citep{yi2017semismooth} and 
Student's loss \citep{mai2025heavy}. 
The simulation results indicate that our method consistently exhibits strong empirical performance relative to these alternatives, particularly in challenging settings with non-Gaussian errors or contaminated by outliers.
Moreover, our proposed method outperforms classical Lasso even in the Gaussian noise.
In addition to simulations, we present real data applications that further supports the utility of our approach. 
Numerical results show that our proposed method are very competitive and promising. 

The remainder of the paper is organized as follows. Section~\ref{sc_model_method} introduces the proposed methodology alongside the Exponential Lasso approach and provides theoretical insights into the robustness properties of the loss function. This section also establishes non-asymptotic statistical guarantees for the proposed estimator. Section~\ref{sc_algorithm} details the algorithmic implementation and includes a convergence analysis of the proposed optimization scheme. Simulation studies evaluating empirical performance are presented in Section~\ref{sc_simulations}. Applications to two real datasets are discussed in Section~\ref{sc_real_data}. Finally, concluding remarks and discussions are provided in Section~\ref{sc_conclusion}, while all technical proofs are collected in Appendix~\ref{sc_proofs}.

\section{Model and method}
\label{sc_model_method}

\subsection{Robust Lasso with Exponential-type loss}

Let \( \{(x_i, y_i)\}_{i=1}^n\) denote a collection of independent and identically distributed (i.i.d) observations arising from the linear model
\begin{equation}
\label{eq_linear_model}
  y_i = x_i^\top \beta^* + \epsilon_i,   
\end{equation}
where \(x_i \in \mathbb{R}^p\) is the \(i\)-th row of the design matrix \(X\) and 
\(\beta^* \in \mathbb{R}^p\) is the unknown vector of regression coefficients that we seek to estimate.
The condition on the random noise is given below in which we consider a very wild class of noise that cover both heavy-tailed noise and outlier contaminated models.

Let
\begin{equation}\label{eq:correntropy_loss_repeat}
L_\tau(\beta) 
= 
\frac{1}{n}\sum_{i=1}^n \ell_\tau(y_i-x_i^\top\beta),
\quad
\text{ where }
\quad
\ell_\tau(r)
=
\frac{1}{\tau}\big(1-e^{-\tfrac{\tau}{2}r^2}\big)
.
\end{equation}
We consider the following robust penalized regression estimator, called Exponential Lasso:
\begin{equation}
\label{eq:correntropy_lasso}
\widehat{\beta}
    :=
    \arg\min_{\beta \in \mathbb{R}^p}
    \left\{
L_\tau(\beta) 
    + \lambda \|\beta\|_1
    \right\},
\end{equation}
where $\tau>0$ controls the degree of robustness and $\lambda>0$ is a regularization parameter.
For small $\tau$, the loss approximates the quadratic loss and~\eqref{eq:correntropy_lasso} reduces to the standard Lasso.
For larger $\tau$, the exponential term strongly downweights large residuals, thus providing robustness against outliers.

\subsection{Robustness of the loss}
\begin{figure}[!h]
    \centering
    \includegraphics[width=14cm]{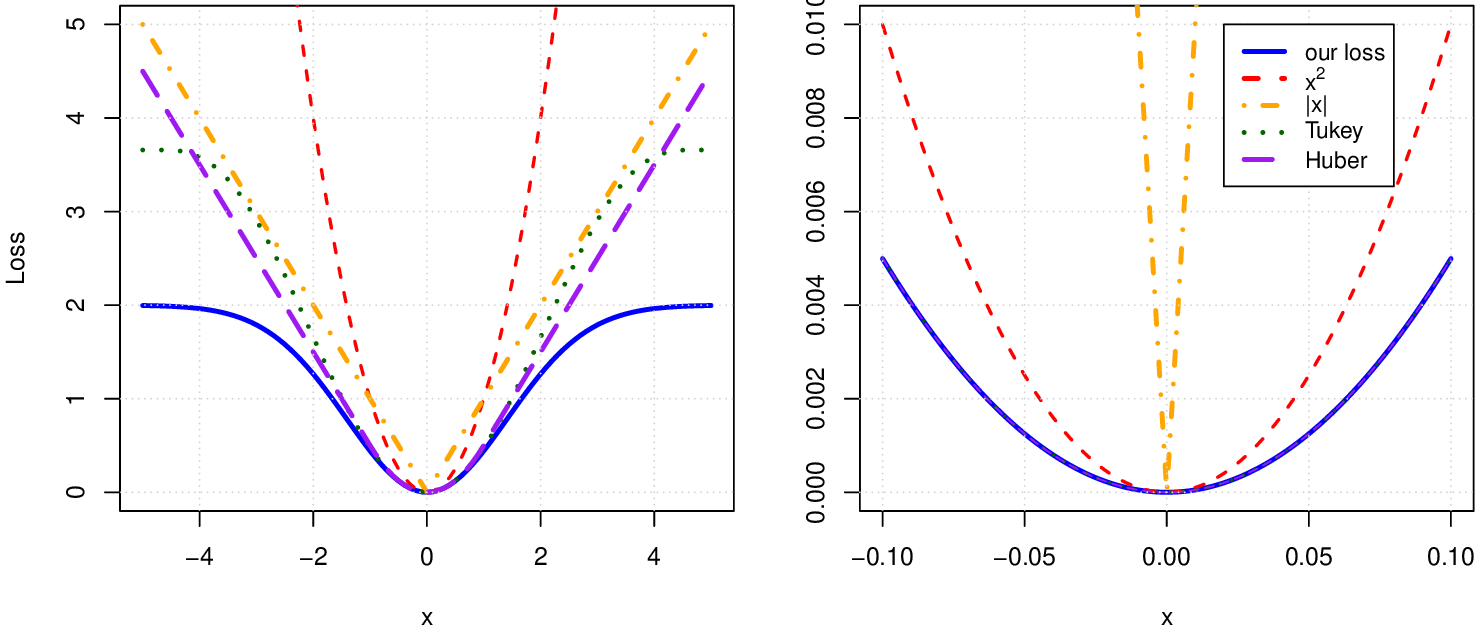}
    \caption{\it Comparison of our loss function ($\tau = 0.5$) with other common losses: squared loss, absolute ($\ell_1$) loss, Tukey’s biweight loss, and Huber loss. The plot illustrates that, unlike Huber loss, our loss is much less sensitive to large residuals while closely resembling the squared loss for small residual values. Left: full-scale plot. Right: zoomed-in view near zero residuals. 
    }
    \label{fig_compare_robustloss}
\end{figure}

\paragraph{Some intuitions:}
One can explicitly show the connection to the standard Lasso. The exponential loss $\ell_\tau(r)$ can be analyzed using a Taylor expansion for $e^x$ around 0. Let $u = -\frac{\tau}{2}r^2$. Since $e^u \approx 1 + u$ for small $u$:
$$  
\ell_\tau(r) 
=
\frac{1}{\tau}\left(1 - e^{-\frac{\tau}{2}r^2}\right) 
\approx 
\frac{1}{\tau}\left(1 - \left(1 - \frac{\tau}{2}r^2\right)\right) 
=
\frac{1}{\tau}\left(\frac{\tau}{2}r^2\right) 
=
\frac{1}{2}r^2
.
$$
This formally demonstrates that as $\tau \to 0$, your objective function $L_\tau(\beta)$ converges to the standard least-squares loss,
and thus $\widehat{\beta}$ converges to the classical Lasso estimator. 
This highlights that our proposed method is a natural generalization of the classical Lasso. See Figure \ref{fig_compare_robustloss} for a detailed visualization comparison between different losses: squared loss, Tukey’s biweight loss, absolute ($\ell_1$) loss,  and Huber loss. The plot illustrates that, unlike Huber loss, our loss is much less sensitive to large residuals while closely resembling the squared loss for small residual values.

\paragraph{The Influence Function:}
A key concept in robust statistics is the influence function \citep{hampel1974influence}, which measures the effect of an infinitesimal outlier on the estimator. The influence function is proportional to the first derivative of the loss function, $\psi(r) = \ell_\tau'(r)$, where
$$  
\psi(r) 
=
\frac{\partial}{\partial r} \left[ \frac{1}{\tau}\left(1-e^{-\tfrac{\tau}{2}r^2}\right) \right] = r e^{-\tfrac{\tau}{2}r^2}
.
$$
\begin{itemize}
    \item The influence function is bounded: The maximum value of $|r \exp{(-\frac{\tau}{2}r^2)}|$ is finite. This contrasts with the L2 loss, where $\psi(r) = r$, which is unbounded. An unbounded influence function means a single large outlier can have an arbitrarily large (i.e., infinite) influence on the estimate.
    \item It is redescending: As the residual $r \to \infty$ (a gross outlier), the influence $\psi(r) \to 0$. This is a very strong form of robustness. The estimator completely ignores data points that are sufficiently far from the bulk of the data. This is an advantage over other robust losses like the Huber loss, whose influence function is bounded but not redescending (it becomes constant, $\psi(r) = \text{sign}(r) \cdot k$, for large $r$).
\end{itemize}

\paragraph{Other insights:}
The proposed loss function
can also be viewed as a correntropy or Welsch-type loss that downweights large residuals through an exponential kernel, thereby enhancing robustness against outliers, which is known in the robust signal-processing  literature \citep{liu2007correntropy,he2011robust} .
Moreover, from an information-theoretic perspective, it is also closely connected to the $\alpha$-divergence, where \(\tau\) acts analogously to the divergence parameter controlling the trade-off between efficiency and robustness \citep{iqbal2019alpha,rekavandi2021robust}.
This dual interpretation highlights the method’s grounding in both robust estimation and divergence-based statistical learning.
We also note that similar exponential-type loss functions have been explored in prior studies as in \citep{wang2013robust,song2021robust,wang2023robust}, where the loss takes the form $ \ell_\tau(t) 
=
1 - e^{-t^2/ \tau }  $. 
In contrast, our proposed formulation is more directly connected to the $\alpha$-divergence family. Furthermore, while the aforementioned works primarily address low-dimensional settings and provide asymptotic analyses, our study focuses on high-dimensional regimes and establishes non-asymptotic theoretical guarantees.

\subsection{Statistical guarantee}
We now demonstrate that, under suitable assumptions, our Exponential-Lasso method achieves strong non-asymptotic theoretical guarantees comparable to those established for the Huber loss.

We make the following assumptions.

\begin{assume}[Design and sparsity]
\label{asm:design}
We assume that:
\begin{itemize}
    \item[a).] The rows \(x_i\in\mathbb R^p\) are non-random or random but satisfy \(\|x_i\|_\infty\le K\) almost surely for a known constant \(K>0\).

    \item[b).]  Let \(\beta^*\) be the true parameter with support \(S=\operatorname{supp}(\beta^*)\) and sparsity \(s:=|S| <n <p\).

    \item[c).] There exists \(\phi_{\min}>0\) and a radius \(r>0\) such that for all \(\Delta\in\mathbb R^p\) with \(\|\Delta\|_2\le r\) and \(\|\Delta_{S^c}\|_1\le 3\|\Delta_S\|_1\),
\[
\frac{1}{n}\sum_{i=1}^n (x_i^\top\Delta)^2 \;\ge\; \phi_{\min}\,\|\Delta\|_2^2.
\]
\end{itemize}

\end{assume}

\begin{assume}[Noise]
\label{asm:noise}
The errors \(\varepsilon_i\) are i.i.d. with the following properties:
\begin{itemize}
  \item[(i).] \(\varepsilon_i\) are symmetric about \(0\).
  \item[(ii).] There exists a constant \(c\in(0,1/\sqrt{\tau})\) and \(p_0:=\mathbb P(|\varepsilon_i|\le c)>0\).
\end{itemize}
\end{assume}

\noindent Remarks: symmetry can be relaxed by replacing the centering step with the population bias, but symmetry keeps statements concise. The choice \(c<1/\sqrt\tau\) ensures positive curvature of the per-observation second derivative inside \(|r|\le c\).

\paragraph{Discussion on noise assumptions.}
The noise conditions assumed above
are notably weaker than the conventional sub-Gaussian or even sub-exponential assumptions often imposed in high-dimensional regression analysis. 
In particular, condition~(ii) only requires that the noise distribution has some probability mass around zero, ensuring that the majority of samples are not dominated by extreme outliers, while symmetry guarantees that the noise has zero median. 
No moment or exponential tail condition is imposed, allowing the framework to accommodate a broad class of heavy-tailed or contamination models, such as Student’s $t$ distributions with small degrees of freedom or Huber error models of the form 
\[
\varepsilon_i \sim (1-\pi)\,N(0,\sigma^2) + \pi\,G,
\]
where $G$ may represent a heavy-tailed or outlier-generating component (e.g., Cauchy or Laplace). 
Therefore, this assumption captures realistic data-generating mechanisms with occasional large deviations, while retaining sufficient regularity for establishing estimation error bounds. 
It is thus particularly suitable for robust (penalized) regression models designed to handle impulsive noise or mild contamination in the observations.

Define the positive curvature on the interval \([-c,c]\),
\[
\underline\gamma := \min_{|u|\le c} e^{-\tfrac{\tau}{2}u^2}(1-\tau u^2) \;=\; e^{-\tfrac{\tau}{2}c^2}(1-\tau c^2) \;>\;0.
\]

\begin{theorem}\label{thm:correntropy_lasso}
Under Assumptions \ref{asm:design}--\ref{asm:noise}, fix \(\delta\in(0,1)\). 
Choose the tuning parameter
\begin{equation}\label{eq:lambda_choice}
\lambda  \;=\; \frac{4K}{\sqrt{e\tau}}\sqrt{\frac{2\log(2p/\delta)}{n}} .
\end{equation}
Assume \(r>0\) in Assumption \ref{asm:design} is small enough so that for all \(\Delta\) in the cone \(\{\|\Delta\|_2\le r,\ \|\Delta_{S^c}\|_1\le 3\|\Delta_S\|_1\}\) it holds that \(|x_i^\top\Delta|\le c/2\) for all \(i\) (this is satisfied when \(r\) is chosen such that \(K\sqrt{s}\, r \le c/2\)). 

Then with probability at least \(1-\delta - 2\exp(-n p_0^2/8)\) 
any global minimizer \(\widehat\beta\) satisfying \(\|\widehat\beta-\beta^*\|_2\le r\) obeys
\begin{equation}\label{eq:thm_rate}
\|\widehat\beta-\beta^*\|_2 
\;\le\;
\frac{12\,\lambda\sqrt{s}}{\kappa}
,
\quad\text{and}\quad
\|\widehat\beta-\beta^*\|_1 
\;\le\; 
\frac{48\,\lambda s}{\kappa}
,
\end{equation}
where the constant \(\kappa>0\) can be taken as
$
\kappa \;=\; \frac{p_0}{2}\,\underline\gamma\,\phi_{\min}.
$
In particular, with the choice \eqref{eq:lambda_choice} we obtain the explicit bound
\[
\|\widehat\beta-\beta^*\|_2 \;\le\; 
\frac{48 K}{\kappa\sqrt{e\tau}}\sqrt{\frac{s\log(2p/\delta)}{n}}.
\]
\end{theorem}

The theoretical analysis is based on the Local Restricted Strong Convexity (LRSC) condition, with a full proof provided in Appendix \ref{sc_proofs}. The methodology largely adopts the general framework proposed in \cite{loh2017statistical,loh2021scale}, and as such, our results parallel those obtained for the Huber-loss Lasso. The primary contribution of our approach is the reliance on a substantially weaker noise condition, as we do not assume the existence of any moments for the noise distribution.

\section{Majorization–Minimization Algorithm}
\label{sc_algorithm}
\subsection{Algorithm development}

The exponential-type loss in~\eqref{eq:correntropy_lasso} is nonconvex but smooth.
To derive an efficient iterative algorithm, we adopt a
majorization–minimization (MM) approach.  
Let us define for each observation
\[
\ell_i(\beta)
=
\frac{1}{\tau}\Big(1 - \exp 
\left(-\frac{\tau}{2} r_i(\beta)^2 \right)\Big),
\quad
r_i(\beta) = y_i - x_i^\top \beta.
\]
Consider the function $\phi(u) = 1 - \exp(-\tfrac{\tau}{2}u)$ for $u \ge 0$.
Since $\phi''(u) = -(\tfrac{\tau}{2})^2 \exp(-\tfrac{\tau}{2}u) < 0$, the function $\phi$ is concave.
For any fixed $u^{(t)}$, the first-order Taylor expansion provides a global upper bound (a tight majorizer):
\begin{equation}
\phi(u)
\le
\phi(u^{(t)}) + \phi'(u^{(t)})(u - u^{(t)}),
\quad \forall\,u\ge 0.
\label{eq:phi-major}
\end{equation}
Substituting $u_i = r_i(\beta)^2$ and summing over $i$ yields the following upper bound for the empirical loss:
\begin{equation}
    \label{eq:surrogate}
    \frac{1}{n}\sum_{i=1}^n \ell_i(\beta)
\le
C + \frac{1}{2n} \sum_{i=1}^n v_i^{(t)} r_i(\beta)^2,
\end{equation}
where $C$ is a constant independent of $\beta$, and
\begin{equation}
v_i^{(t)} = \exp\!\left(-\frac{\tau}{2} r_i(\beta^{(t)})^2\right)
\label{eq:weight_update}
\end{equation}
acts as an adaptive weight.
Minimizing the right-hand side of the above inequality thus defines the MM update.

Given the current estimate $\beta^{(t)}$, the MM procedure alternates between
updating the weights $v_i^{(t)}$ according to~\eqref{eq:weight_update} and
solving a weighted Lasso subproblem:

\begin{itemize}
\item \textbf{(Step 1)}
Compute residuals $r_i^{(t)} = y_i - x_i^\top \beta^{(t)}$ and update the weights
\[
v_i^{(t)} = \exp\!\left(-\frac{\tau}{2} (r_i^{(t)})^2\right),
\qquad i = 1, \dots, n.
\]
These weights downweight observations with large residuals, thereby reducing the
influence of outliers.

\item \textbf{(Step 2)}
Update the regression coefficients by solving the weighted Lasso problem
\begin{equation}
\label{eq:weighted_lasso}
\beta^{(t+1)} 
=
\arg\min_{\beta \in \mathbb{R}^p} Q^{(t)}(\beta)
:=
\arg\min_{\beta \in \mathbb{R}^p}
\left\{
\frac{1}{2n}\sum_{i=1}^n v_i^{(t)} (y_i - x_i^\top \beta)^2
+ \lambda \|\beta\|_1
\right\}.
\end{equation}
\end{itemize}

The above two steps are repeated until convergence, e.g.,
until $\|\beta^{(t+1)} - \beta^{(t)}\|_2 / (1 + \|\beta^{(t)}\|_2) < \varepsilon$ for a small tolerance $\varepsilon > 0$.
Step 2 can be efficiently implemented using standard coordinate-descent
algorithms or existing Lasso solvers (e.g., \texttt{glmnet} in \textsf{R})
by specifying the observation weights $\{v_i^{(t)}\}$. The outline of our proposed algorithm given in Algorithm~\ref{algorithm_main}.

\subsection{Coordinate descent updates using soft-thresholding}

We estimate the regression coefficients in \eqref{eq:weighted_lasso} via a coordinate descent algorithm. Let denote the residual vector at iteration $t $ as
$ \quad
r^{(t)} = y - X \beta^{(t)}
$
.
For the \(j\)-th coordinate, we define the corresponding partial residual — that is, the residual excluding the contribution of variable \(j\) — as
\[
r_j = r^{(t)} + X_j \beta_j^{(t)}.
\]
The objective function restricted to the single coefficient \(\beta_j\) can then be expressed as:
\[
\frac{1}{2n} \sum_{i=1}^n v_i (r_{ij} - x_{ij} \beta_j)^2 + \lambda |\beta_j|.
\]
Minimization of this univariate problem yields a closed-form update via the soft-thresholding operator, \citep{donoho1994ideal}:
\[
z_j := \sum_{i=1}^n v_i x_{ij} r_{ij}, \quad
U_j := \sum_{i=1}^n v_i x_{ij}^2, 
\quad
\beta_j \leftarrow \frac{1}{U_j} 
\mathcal{S}(z_j, \lambda),
\]
where 
$$
\mathcal{S}
(z, \lambda) = \text{sign}(z) \cdot \max(|z| - \lambda, 0)
,
$$
denotes the soft-thresholding function.

\paragraph{Discussion.}
The proposed algorithm can be viewed as an EM-like procedure, where
the weights $\{v_i^{(t)}\}$ play a role analogous to latent variables that reflect
the reliability of each observation.
As $\tau \to 0$, all $v_i^{(t)} \to 1$, and the algorithm reduces to the ordinary
Lasso.
For larger $\tau$, the exponential decay of $v_i^{(t)}$
produces strong robustness to large residuals.

\begin{algorithm}[H]
\caption{MM Algorithm for Exponential-type robust Lasso}
\begin{algorithmic}[1]
\STATE \textbf{Input:} data $(x_i,y_i)_{i=1}^n$, tuning parameters $\tau>0$, $\lambda \geq 0$.
\STATE Initialize $\beta^{(0)}$ (e.g., by ordinary Lasso).
\REPEAT
\STATE Compute residuals $r_i^{(t)} = y_i - x_i^\top \beta^{(t)}$.
\STATE Update weights $v_i^{(t)} = \exp\!\left(-\frac{\tau}{2} (r_i^{(t)})^2\right)$.
\STATE Solve weighted Lasso:
\[
\beta^{(t+1)} =
\arg\min_{\beta}
\left\{
\frac{1}{2n}\sum_{i=1}^n v_i^{(t)} (y_i - x_i^\top \beta)^2
+ \lambda \|\beta\|_1
\right\}.
\]
\UNTIL{convergence criterion is met.}
\STATE \textbf{Output:} final estimate $\widehat{\beta} = \beta^{(t+1)}$.
\end{algorithmic}
\label{algorithm_main}
\end{algorithm}

Under standard regularity conditions for MM algorithms, the sequence
$\{\beta^{(t)}\}$ monotonically decreases the objective in~\eqref{eq:correntropy_lasso}
and converges to a stationary point.

\subsection{Convergence analysis of the MM algorithm}
\label{sec:convergence}

We analyze the Majorization–Minimization (MM) algorithm given in Algorithm~\ref{algorithm_main} for the exponential-type robust Lasso objective
\begin{equation}
\label{eq:obj}
F(\beta)
=
\frac{1}{n}\sum_{i=1}^n \ell_i(\beta) + \lambda \|\beta\|_1
\quad\text{with}
\quad
\ell_i(\beta)
=
\frac{1}{\tau}\Big(1 - \exp\big(-\frac{\tau}{2} r_i(\beta)^2\big)\Big),
\quad 
r_i(\beta)=y_i-x_i^\top\beta,
\end{equation}
where $\tau>0$ and $\lambda\ge 0$. For convenience write the smooth (nonconvex) loss part as
\[
L(\beta) := \frac{1}{n}\sum_{i=1}^n \ell_i(\beta),
\]
so $F(\beta)=L(\beta)+\lambda\|\beta\|_1$.

We first show that the MM iterates produce a monotone decreasing sequence of objective values and remain in a bounded level set.

\begin{theorem}[Monotone decrease and boundedness]
\label{thm:descent}
Assume that each row $x_i$ satisfies $\|x_i\|_2<\infty$. Let $\{\beta^{(t)}\}$ be the sequence produced by Algorithm~\ref{algorithm_main} where in each M-step we compute an exact minimizer of the surrogate $Q^{(t)}(\beta)$ in \eqref{eq:weighted_lasso}. Then:
\begin{enumerate}
  \item[(i).] (Descent) The sequence of objective values $\{F(\beta^{(t)})\}$ is nonincreasing:
  $
  F(\beta^{(t+1)}) \le F(\beta^{(t)}),  \forall t\ge 0.
  $
  \item[(ii).] (Lower bounded) $F(\beta)\ge 0$ for all $\beta$, hence $\{F(\beta^{(t)})\}$ converges to a finite limit $F^\star$.
  \item[(iii).] (Bounded iterates) The iterates $\{\beta^{(t)}\}$ lie in the sublevel set $\{\beta: F(\beta)\le F(\beta^{(0)})\}$, which is bounded under the mild condition that $\lambda>0$ or that $X$ has full column rank; hence $\{\beta^{(t)}\}$ is bounded.
\end{enumerate}
\end{theorem}

Next we establish that any accumulation point of the MM iterates is a stationary point of the nonconvex objective $F(\beta)$. Because $F$ contains the nondifferentiable $\ell_1$ term, stationarity is understood in the subgradient/KKT sense.

\begin{theorem}[Cluster points are stationary]
\label{thm:stationary}
Let assume that each row $x_i$ satisfies $\|x_i\|_2<\infty$ and let $\{\beta^{(t)}\}$ be generated by Algorithm~\ref{algorithm_main} with exact M-steps (exact minimizer of the surrogate). Then every limit point $\beta^\star$ of $\{\beta^{(t)}\}$ is a stationary point of $F$, i.e.
$ 0 \in \nabla L(\beta^\star) + \lambda\partial\|\beta^\star\|_1.$
Consequently, the whole sequence has its set of cluster points contained in the set of stationary points of $F$.
\end{theorem}

\begin{corollary}[Convergence of objective and subsequential stationarity]
Under the hypotheses of Theorems~\ref{thm:descent}--\ref{thm:stationary}, the objective values $F(\beta^{(t)})$ converge to a finite $F^\star$ and every cluster point of $\{\beta^{(t)}\}$ is a stationary point of $F$. If, in addition, the set of stationary points at level $F^\star$ is finite and the sequence has a unique cluster point, then $\beta^{(t)}\to\beta^\star$.
\end{corollary}

The objective $F(\beta)$ is nonconvex because $L(\beta)$ is nonconvex; therefore MM can only be expected to converge to a local stationary point in general. 
Uniqueness of the weighted Lasso minimizer in each M-step holds when $X^\top \tilde V^{(t)} X$ is positive definite (e.g. full column rank and all $\tilde v_i^{(t)}>0$). In practice $\tilde v_i^{(t)}\in(0,1]$, so strict positive definiteness reduces to conditions on $X$.

\subsection{Tuning the regularization parameter $\lambda$}

We select the regularization parameter 
\(\lambda\) in the exponential Lasso method primarily using \(K\)-fold cross-validation (CV), which serves as the default option. Cross-validation evaluates predictive performance by partitioning the data into training and validation folds and choosing the 
\(\lambda\) that minimizes the average prediction error. This provides a practical, data-driven balance between model sparsity and predictive accuracy.

Our method is implemented in the \texttt{R} package \texttt{heavylasso} available on Github: \url{https://github.com/tienmt/heavylasso} .

\section{Simulation studies}
\label{sc_simulations}

\subsection{Setup}
\subsubsection*{Compared methods}
The central aim of this simulation is to evaluate how different robust loss functions affect the performance of the Lasso estimator. Accordingly, our investigation is strictly focused on regression methods that incorporate the Lasso penalty, to the exclusion of other regularization approaches.

The following four estimators are included in our comparative analysis:
\begin{itemize}
    \item Classical Lasso, which minimizes a squared loss function, implemented in the \texttt{R} package \texttt{glmnet} \citep{glmnetpackage}.

\item Huber Lasso, which employs the hybrid Huber loss function.
        
    \item LAD Lasso,  which is based on  the $\ell_1$ loss function. Both the LAD and Huber variants are implemented in the \texttt{R} package \texttt{hqreg}.
    
    \item Heavy Lasso, which utilizes a Student loss function, available in the \texttt{R} package \texttt{heavylasso} \citep{mai2025heavy}.
\end{itemize}

\subsubsection*{Simulation settings}
\noindent
In our data generation process, the predictors $X_i$ follow a $N(0, \Sigma)$ distribution. We investigate two specific covariance scenarios: (i) an identity matrix ($\Sigma = \mathbb{I}_p$), representing independent predictors, and (ii) an autoregressive correlation structure ($\Sigma_{ij} = \rho_X^{|i-j|}$). 
\\
The ground-truth vector $\beta_0$ is $s$-sparse, with the $s$ non-zero coefficients split evenly, taking values of $1$ (for $s/2$ entries) and $-1$ (for the remaining $s/2$ entries). The responses $y_i$ are then produced via the linear model \eqref{eq_linear_model} under the following noise conditions:
\begin{itemize}
    \item Gaussian noise, $ \epsilon_i \sim \mathcal{N} (0,1) $. This serves as a baseline to assess how various robust methods perform under ideal (light-tailed) conditions.
    
    \item Gaussian noise with large variance. $ \epsilon_i \sim \mathcal{N} (0,3^2) $.
    This setting introduces moderate heavy-tailed behavior through increased variance.
    
    \item Student noise. $ \epsilon_i \sim t_3 $. This case represents heavy-tailed noise with finite variance.
    
    \item Cauchy noise. $ \epsilon_i \sim Cauchy $. This represents a more extreme heavy-tailed setting with infinite variance. 

\item Contaminated with outliers. $ \epsilon_i \sim \mathcal{N} (0,1) $ or $ \epsilon_i \sim t_3 $  but some portion of the observed responses are further contaminated by outliers. This setting evaluates robustness to contamination.
\end{itemize}

We assessed the performance of each method from three different angles: parameter estimation, prediction on new data, and variable selection accuracy.
We used two metrics to evaluate how accurately each model estimated the true coefficients. First, we measured the estimation error using the squared $\ell_2$ norm, which calculates the distance between the estimated coefficients ($\widehat{\beta}$) and the true coefficients ($\beta_0$):
$ \:\: \| \widehat{\beta} - \beta_0 \|_2^2$.
Second, we evaluated the error in the model's fitted values on the training data, captured by the linear predictor error:
$$\ell(X^\top\beta_0 ) := \frac{1}{n} \| X^\top (\widehat{\beta} - \beta_0) \|_2^2$$
To measure prediction accuracy, we calculated the mean squared prediction error (MSPE) on a large, independent test set. This test set, $(X_{\text{test}}, y_{\text{test}})$, was generated from the same model as the training data, with a fixed size of $n_{\text{test}} = 5000$. The MSPE is defined as:
$$
\text{MSPE}_{\text{test}} := \frac{1}{n_{\text{test}}} \sum_{i=1}^{n_{\text{test}}} \left( y_{{\text{test}},i} - X_{{\text{test}},i}^\top \widehat{\beta} \right)^2
.
$$
Finally, we assessed each method's ability to correctly identify the relevant predictors in the model. This was measured using two standard metrics: the True Positive Rate (TPR) and the False Discovery Rate (FDR).

Each simulation setting is repeated 100 times, and we report the mean and standard deviation of the results. The outcomes are presented in Tables \ref{tb_low_dim}, \ref{tb_low_dim_rX}, \ref{tb_high_dim}, 
\ref{tb_high_dim_rX} and \ref{tb_outlier}.
The regularized parameters for all five methods are selected via 5-fold cross-validation. The tuning parameter $\tau $ in our method is set to $0.1$, which is motivated from sensitivity analysis in Subsection \ref{sc_sub_sensitivity} below.

\subsection{Simulations results}

\subsubsection{Results with Heavy-tailed noises}

We first examine the performance of our method against competing approaches across several noise settings. This comparison is conducted in two distinct regimes: a small scale setting with $p=120$ and $n=100$ (Tables \ref{tb_low_dim}, \ref{tb_low_dim_rX}), and a medium scale setting with $p=500$ and $n=300$ (Tables \ref{tb_high_dim}, \ref{tb_high_dim_rX}). The true sparsity $s^*$ is fixed at $10$ for all experiments.

In both settings, particularly under Gaussian noise, our proposed method and the Heavy Lasso are shown to be highly competitive. They frequently outperform the classical Lasso, a level of performance not attained by the Huber Lasso or the L1 Lasso in these experiments. The robustness and superiority of these results appear to improve in the larger-scale data scenarios, as evidenced in Tables \ref{tb_high_dim} and \ref{tb_high_dim_rX}.

A general observation is that our proposed method and the Heavy Lasso, both being non-convex, tend to return more small non-zero coefficients. This characteristic typically leads to a higher false positive rate (or false discovery rate) when compared to convex alternatives like the Huber Lasso or the L1 Lasso. Despite this, a crucial advantage is their ability to consistently select the true support.

When considering heavy-tailed noise, such as the Student's $t_3$ or Cauchy distributions, our proposed method and the Heavy Lasso invariably provide the best results regarding both estimation and prediction errors. More particularly, in the challenging Cauchy noise scenario, our method demonstrates clear superiority across all metrics, including estimation, prediction, and variable selection accuracy.

\subsubsection{Results with Outliers}

We next evaluate the behavior of all considered methods in the presence of outliers. For this purpose, we fix the simulation parameters at $p=500$, $s^*=10$, and $n=300$. We investigate two underlying noise distributions: standard normal, $\mathcal{N}(0,1)$, and Student's $t_3$. The proportion of the response data contaminated by outliers is varied among 10\%, 20\%, and 30\%. The results of this analysis are presented in Table \ref{tb_outlier}.

The results indicate that our proposed method outperforms both the Huber Lasso and the L1 Lasso. As expected, the classical Lasso breaks down entirely under these conditions. When the underlying noise is Gaussian, our method is significantly better than the Heavy Lasso (which utilizes a Student's loss). This performance gap is particularly pronounced when a larger fraction of the responses are contaminated, for example, at the 30\% level. In the presence of Student's $t_3$ noise, the Heavy Lasso exhibits a slight advantage at lower contamination levels. However, as the contamination fraction increases to 30\%, our proposed method once again outperforms the Heavy Lasso. These findings clearly highlight the superior robustness of our proposed methodology.

\begin{table}[!ht]
\centering
\caption{\small \it Simulation for various loss functions with Lasso penalization, under the setting $ p = 120, s^* = 10, n = 100 $ and independent predictors. The reported values are the mean across 100 simulation repetitions, with the standard deviation provided in parentheses. Bold font highlights the superior method. TPR: true positive rate; FDR: false discovery rate; $\text{MSPE}_{\text{test}}$: mean squared prediction error on testing data. }
\small
\begin{tabular}{ l l | cccc cc }
		\hline \hline
Noise & Method (loss) 
& $ \| \widehat{\beta} -\beta_0 \|_2^2 $ 
& $ \ell(X^\top\beta_0 ) $ 
& $\text{MSPE}_{\text{test}}$
& TPR
& FPR
\\
\hline \hline
$ \mathcal{N} (0,1) $  
& proposed loss
& \textbf{0.62} (0.22) & \textbf{0.40} (0.11) & \textbf{1.62} (0.22) 
& \textbf{1.00} (0.00) &   0.75 (0.06) 
\\
& Student's loss
& 0.64 (0.26) & \textbf{0.40} (0.11) & 1.64 (0.26)
& \textbf{1.00} (0.00) & 0.75 (0.06)
\\
& squared   loss
& 0.78 (0.26) &  0.55 (0.18) & 1.82 (0.26)
& \textbf{1.00} (0.00) & 0.45 (0.13)
\\
& $\ell_1$   loss
& 1.09 (0.43) & 0.71 (0.24) & 2.02 (0.43) 
& \textbf{1.00} (0.00) &   0.65 (0.10)
\\
& Huber  loss
& 0.95 (0.41) & 0.65 (0.27) & 1.93 (0.40) 
& \textbf{1.00} (0.00) &   0.57 (0.10)
\\
		\hline
		\hline	
$ \mathcal{N} (0,3) $  
& proposed loss
& 7.06 (2.01) & 5.34 (1.77) & 16.0 (2.07) 
& 0.78 (0.28) & 0.78 (0.13)
\\
& Student's loss
& \textbf{5.24} (1.67) & \textbf{3.79} (1.17) &  \textbf{14.3} (1.74)
&  \textbf{0.89} (0.17) & 0.72 (0.10)
\\
& squared   loss
& 7.00 (2.26) & 6.31 (2.52) &  16.0 (2.32)
& 0.52 (0.36) & 0.25 (0.23)
\\
& $\ell_1$   loss
& 8.29 (1.91) & 7.82 (2.50) & 17.3 (1.96)
& 0.32 (0.33) & 0.22 (0.28)
\\
& Huber  loss
& 7.89 (1.94) & 7.23 (2.41) & 16.9  (1.99)
& 0.40 (0.33) &  0.26 (0.27)
\\
		\hline
		\hline	
$ t_3 $ 
& proposed loss
& \textbf{1.20} (0.52) & \textbf{0.77} (0.37) & \textbf{4.15} (0.78) 
& \textbf{1.00} (0.00) &  0.76 (0.06)
\\
& Student's loss
& 1.29 (0.61) & 0.81 (0.32) & 4.25 (0.78) 
& \textbf{1.00} (0.00) &  0.75 (0.06)
\\
& squared   loss
& 2.72 (2.29) & 2.10 (2.06) &  5.67 (2.27)
& 0.93 (0.24) & 0.37 (0.17) 
\\
& $\ell_1$   loss
& 1.87 (1.05) & 1.27 (0.70) & 5.03 (1.14) 
& 0.99 (0.04) &  0.61 (0.12)
\\
& Huber  loss
& 1.73 (0.87) & 1.20 (0.63) &  4.71 (1.03)
& 0.99 (0.04) & 0.53 (0.11) 
\\
		\hline
		\hline	
$ Cauchy $ 
& proposed loss
& \textbf{5.17} (3.23) & \textbf{3.71} (2.10) &  $>\!\! 10^5$
& 0.89 (0.25) &  0.71 (0.21)
\\
& Student's loss
& 5.82 (5.19) & 4.01 (4.33) &   $>\!\! 10^5$
& \textbf{0.91} (0.24) & 0.71 (0.25) 
\\
& squared   loss
& 9.97 (0.14) & 9.95 (1.33) &   $>\!\! 10^5$
& 0.01 (0.06) &  \textbf{0.01} (0.11)
\\
& $\ell_1$   loss
& 9.19 (2.11) & 8.57 (2.72) &   $>\!\! 10^5$
& 0.13 (0.31) &   0.13 (0.24)
\\
& Huber  loss
& 9.13 (2.14) & 8.59 (2.71) &   $>\!\! 10^5$
& 0.15 (0.33) &  0.10 (0.18)
\\
		\hline
		\hline	
\end{tabular}
\label{tb_low_dim}
\end{table}

\begin{table}[!ht]
\centering
\caption{\small \it Simulation for various loss functions with Lasso penalization, under the setting $ p = 120, s^* = 10, n = 100 $ and correlated design $\rho_X = 0.5$. The reported values are the mean across 100 simulation repetitions, with the standard deviation provided in parentheses. Bold font highlights the superior method. TPR: true positive rate; FDR: false discovery rate; $\text{MSPE}_{\text{test}}$: mean squared prediction error on testing data.  }
\small
\begin{tabular}{ l l | cccc cc }
		\hline \hline
Noise & Method (loss) 
& $ \| \widehat{\beta} -\beta_0 \|_2^2 $ 
& $ \ell(X^\top\beta_0 ) $ 
& $\text{MSPE}_{\text{test}}$
& TPR
& FPR
\\
\hline \hline
$ \mathcal{N} (0,1) $  
& proposed loss
& 0.52 (0.27) & 0.30 (0.09) & 1.42 (0.20)
& \textbf{1.00} (0.00) &   0.70 (0.08)
\\
& Student's loss
& \textbf{0.50} (0.22) & \textbf{0.29} (0.11) & \textbf{1.41} (0.22) 
& \textbf{1.00} (0.00) &   0.75 (0.06) 
\\
& squared   loss
& 0.72 (0.26) &  0.45 (0.18) & 1.59 (0.26)
& \textbf{1.00} (0.00) & \textbf{0.31} (0.13)
\\
& $\ell_1$   loss
& 0.94 (0.43) & 0.59 (0.24) & 1.77 (0.43) 
& \textbf{1.00} (0.00) &   0.48 (0.10)
\\
& Huber  loss
& 0.79 (0.41) & 0.50 (0.27) & 1.65 (0.40) 
& \textbf{1.00} (0.00) &   0.38 (0.10)
\\
		\hline
		\hline	
$ \mathcal{N} (0,3) $  
& proposed loss
& 5.17 (1.85) & 3.67 (1.16) & 13.7 (1.86) 
& 0.91 (0.09) & 0.72 (0.06)
\\
& Student's loss
& \textbf{3.96} (1.67) & \textbf{2.61} (1.17) &  \textbf{12.4} (1.74)
&  \textbf{0.92} (0.10) & 0.66 (0.10)
\\
& squared   loss
& 4.55 (2.26) & 4.31 (2.52) &  14.0 (2.32)
& 0.76 (0.36) & \textbf{0.15} (0.23)
\\
& $\ell_1$   loss
& 5.37 (1.91) & 5.43 (2.50) & 15.3 (1.96)
& 0.67 (0.33) & 0.24 (0.28)
\\
& Huber  loss
& 5.25 (1.94) & 5.29 (2.41) & 15.1  (1.99)
& 0.69 (0.33) &  0.21 (0.27)
\\
		\hline
		\hline	
$ t_3 $ 
& proposed loss
& \textbf{0.98} (0.40) & \textbf{0.56} (0.19) & \textbf{3.91} (1.14)
& 1.00 (0.00) & 0.67 (0.10)
\\
& Student's loss
& 1.03 (0.46) & 0.59 (0.21) & 3.95 (1.16)
& \textbf{1.00} (0.00) &  0.68 (0.10)
\\
& squared   loss
& 2.16 (2.29) & 1.75 (2.16) &  5.28 (2.27)
& 0.93 (0.24) & 0.24 (0.17) 
\\
& $\ell_1$   loss
& 1.83 (1.05) & 1.09 (0.70) & 4.61 (1.14) 
& 0.96 (0.04) &  0.43 (0.12)
\\
& Huber  loss
& 1.65 (0.87) & 1.06 (0.63) &  4.51 (1.03)
& 0.97 (0.04) & 0.34 (0.11) 
\\
		\hline
		\hline	
$ Cauchy $ 
& proposed loss
& \textbf{3.34} (2.71) & \textbf{2.26} (2.10) &  $>\!\! 10^5$
& \textbf{0.96} (0.11) &  0.60 (0.21)
\\
& Student's loss
& 4.48  (4.72) & 2.73 (2.21) &   $>\!\! 10^5$
& 0.94 (0.24) & 0.62 (0.25) 
\\
& squared   loss
& 9.80 (0.14) & 9.95 (1.33) &   $>\!\! 10^5$
& 0.03 (0.06) &  \textbf{0.00} (0.11)
\\
& $\ell_1$   loss
& 7.90 (2.11) & 8.57 (2.72) &   $>\!\! 10^5$
& 0.34 (0.31) &   0.03 (0.24)
\\
& Huber  loss
& 7.71 (2.14) & 8.59 (2.71) &   $>\!\! 10^5$
& 0.35 (0.33) &  0.03 (0.18)
\\
		\hline
		\hline	
\end{tabular}
\label{tb_low_dim_rX}
\end{table}

\begin{table}[!ht]
\centering
\caption{ \small \it Simulation results for various loss functions with Lasso penalization, under the setting $ p = 500, s^* = 10, n = 300 $ and independent predictors.  The reported values are the mean across 100 simulation repetitions, with the standard deviation provided in parentheses. Bold font highlights the superior method. TPR: true positive rate; FDR: false discovery rate; $\text{MSPE}_{\text{test}}$: mean squared prediction error on testing data.   }
\small
\begin{tabular}{ l l | cccc cc }
		\hline \hline
Noise & Method (loss) 
& $ \| \widehat{\beta} -\beta_0 \|_2^2 $ 
& $ \ell(X^\top\beta_0 ) $ 
& $\text{MSPE}_{\text{test}}$
& TPR
& FPR
\\
\hline \hline
$ \mathcal{N} (0,1) $  
& proposed loss
& \textbf{0.23} (0.08) & \textbf{0.19} (0.06) & \textbf{1.23} (0.08) & 1.00 (0.00) & 0.85 (0.04) 
\\
& Student's loss
& \textbf{0.23} (0.08) & \textbf{0.19} (0.06) & \textbf{1.23} (0.08) & 1.00 (0.00) & 0.85 (0.04) 
\\
& squared   loss
& 0.31 (0.10) & 0.28 (0.08) & 1.31 (0.10) & 1.00 (0.00) & \textbf{0.42} (0.17) 
\\
& $\ell_1$   loss
& 0.44 (0.16) & 0.38 (0.12) & 1.44 (0.16) & 1.00 (0.00) & 0.54 (0.16) 
\\
& Huber  loss
& 0.39 (0.14) & 0.35 (0.10) & 1.39 (0.14) & 1.00 (0.00) & 0.48 (0.19) 
\\
		\hline
		\hline	
$ \mathcal{N} (0,3) $  
& proposed loss
& 3.49 (1.32) & 3.00 (1.11) & 12.4 (1.46) & \textbf{1.00} (0.01) & 0.89 (0.04) 
\\
& Student's loss
& \textbf{2.36} (0.59) & \textbf{2.04} (0.51) & \textbf{11.3} (0.67) & \textbf{1.00} (0.00) & 0.87 (0.03) 
\\
& squared   loss
& 2.95 (0.83) & 2.72 (0.80) & 11.9 (0.84) & 0.99 (0.04) & \textbf{0.35} (0.18) 
\\
& $\ell_1$   loss
& 4.41 (1.38) & 4.05 (1.31) & 13.3 (1.44) & 0.91 (0.12) & 0.44 (0.21) 
\\
& Huber  loss
& 4.09 (1.19) & 3.74 (1.15) & 13.0 (1.25) & 0.94 (0.10) & 0.44 (0.19)
\\
		\hline
		\hline	
$ t_3 $ 
& proposed loss
& 0.41 (0.14) & 0.34 (0.11) & 3.50 (0.66) & 1.00 (0.00) & 0.83 (0.05) 
\\
& Student's loss
& \textbf{0.40} (0.12) & \textbf{0.33} (0.09) & \textbf{3.48} (0.66) & 1.00 (0.00) & 0.83 (0.04) 
\\
& squared   loss
& 1.13 (0.63) & 1.05 (0.62) & 4.21 (1.00) & 1.00 (0.00) & \textbf{0.23} (0.16) 
\\
& $\ell_1$   loss
& 0.66 (0.18) & 0.59 (0.16) & 3.74 (0.67) & 1.00 (0.00) & 0.47 (0.19)
\\
& Huber  loss
& 0.64 (0.19) & 0.58 (0.16) & 3.73 (0.68) & 1.00 (0.00) & 0.38 (0.17) 
\\
		\hline
		\hline	
$ Cauchy $ 
& proposed loss
& \textbf{2.56} (2.33) & \textbf{2.17} (1.92) & $>\!\! 10^5$ & \textbf{1.00} (0.00) & 0.82 (0.20) 
\\
& Student's loss
& 2.79 (4.03) & 2.31 (3.16) & $>\!\! 10^5$ & \textbf{1.00} (0.00) & 0.81 (0.15) 
\\
& squared   loss
& 10.0 (0.00) & 10.0 (0.73) & $>\!\! 10^5$ & 0.00 (0.00) & \textbf{0.00} (0.00) 
\\
& $\ell_1$   loss
& 7.97 (3.00) & 7.85 (3.08) & $>\!\! 10^5$ & 0.33 (0.44) & 0.03 (0.08) 
\\
& Huber  loss
& 8.02 (2.92) & 7.89 (3.00) & $>\!\! 10^5$ & 0.33 (0.43) & 0.02 (0.07) 
\\
		\hline
		\hline	
\end{tabular}
\label{tb_high_dim}
\end{table}

\begin{table}[!ht]
\centering
\caption{\small \it Simulation results for various loss functions with Lasso penalization, under the setting $ p = 500, s^* = 10, n = 300 $ and correlated design $\rho_X = 0.5$.  The reported values are the mean across 100 simulation repetitions, with the standard deviation provided in parentheses. Bold font highlights the superior method. TPR: true positive rate; FDR: false discovery rate; $\text{MSPE}_{\text{test}}$: mean squared prediction error on testing data.  }
\small
\begin{tabular}{ l l | cccc cc }
		\hline \hline
Noise & Method (loss) 
& $ \| \widehat{\beta} -\beta_0 \|_2^2 $ 
& $ \ell(X^\top\beta_0 ) $ 
& $\text{MSPE}_{\text{test}}$
& TPR
& FPR
\\
\hline \hline
$ \mathcal{N} (0,1) $  
& proposed loss
& \textbf{0.20} (0.08) & \textbf{0.15} (0.04) & \textbf{1.17} (0.06) & 1.00 (0.00) & 0.80 (0.04) 
\\
& Student's loss
& 0.21 (0.08) & \textbf{0.15} (0.05) & \textbf{1.17} (0.07) & 1.00 (0.00) & 0.80 (0.05) 
\\
& squared   loss
& 0.31 (0.13) & 0.24 (0.07) & 1.26 (0.09) & 1.00 (0.00) & \textbf{0.22} (0.16) 
\\
& $\ell_1$   loss
& 0.44 (0.16) & 0.32 (0.09) & 1.35 (0.10) & 1.00 (0.00) & 0.35 (0.16) 
\\
& Huber  loss
& 0.38 (0.15) & 0.28 (0.08) & 1.31 (0.10) & 1.00 (0.00) & 0.28 (0.16) 
\\
		\hline
		\hline	
$ \mathcal{N} (0,3) $  
& proposed loss
& 2.47 (0.86) & 2.08 (0.88) & 11.2 (1.05) & 0.99 (0.03) & 0.84 (0.06) 
\\
& Student's loss
& \textbf{1.83} (0.63) & \textbf{1.48} (0.57) & \textbf{10.6} (0.68) & \textbf{1.00} (0.00) & 0.82 (0.05) 
\\
& squared   loss
& 2.48 (0.84) & 2.18 (0.83) & 11.3 (1.00) & 0.92 (0.08) & \textbf{0.17} (0.16) 
\\
& $\ell_1$   loss
& 3.00 (0.97) & 2.77 (1.07) & 11.9 (1.19) & 0.89 (0.09) & 0.27 (0.20) 
\\
& Huber  loss
& 2.92 (0.93) & 2.65 (0.99) & 11.8 (1.16) & 0.89 (0.09) & 0.25 (0.18) 
\\
		\hline
		\hline	
$ t_3 $ 
& proposed loss
& \textbf{0.34} (0.12) & \textbf{0.25} (0.08) & \textbf{3.16} (0.31) & 1.00 (0.00) & 0.79 (0.06) 
\\
& Student's loss
& 0.35 (0.14) & 0.27 (0.12) & 3.18 (0.32) & 1.00 (0.00) & 0.79 (0.06) 
\\
& squared   loss
& 1.22 (0.93) & 1.08 (1.20) & 4.04 (1.27) & 0.98 (0.08) & \textbf{0.10} (0.15)
\\
& $\ell_1$   loss
& 0.63 (0.23) & 0.51 (0.15) & 3.43 (0.34) & 1.00 (0.00) & 0.25 (0.15) 
\\
& Huber  loss
& 0.58 (0.22) & 0.46 (0.15) & 3.38 (0.32) & 1.00 (0.00) & 0.20 (0.16) 
\\
		\hline
		\hline	
$ Cauchy $ 
& proposed loss
& \textbf{2.85} (3.41) & \textbf{1.96} (2.13) & $>\!\! 10^6$ & \textbf{1.00} (0.00) & 0.83 (0.16) 
\\
& Student's loss
& 4.64 (8.29) & 2.84 (4.43) & $>\!\! 10^6$  & \textbf{1.00} (0.00) & 0.82 (0.15) 
\\
& squared   loss
& 9.92 (0.53) & 18.3 (1.88) & $>\!\! 10^6$  & 0.01 (0.10) & \textbf{0.00} (0.00) 
\\
& $\ell_1$   loss
& 6.33 (2.89) & 10.1 (6.68) & $>\!\! 10^6$  & 0.52 (0.38) & 0.01 (0.05)
\\
& Huber  loss
& 6.39 (2.90) & 10.2 (6.69) & $>\!\! 10^6$  & 0.52 (0.38) & 0.01 (0.05) 
\\
		\hline
		\hline	
\end{tabular}
\label{tb_high_dim_rX}
\end{table}

\begin{table}[!ht]
\centering
\caption{\small \it Simulation results for various loss functions with Lasso penalization, under the setting $ p = 500, s^* = 10, n = 300 $ and independent predictors. The outliers are increased by 10\%, 20\% and 30\%.
 The reported values are the mean across 100 simulation repetitions, with the standard deviation provided in parentheses.
 Bold font highlights the best method. TPR: true positive rate; FDR: false discovery rate; $\text{MSPE}_{\text{test}}$: mean squared prediction error on testing data.  }
\small
\begin{tabular}{ l l | cccc cc }
		\hline \hline
Noise & Method (loss) 
& $ \| \widehat{\beta} -\beta_0 \|_2^2 $ 
& $ \ell(X^\top\beta_0 ) $ 
& $\text{MSPE}_{\text{test}}$
& TPR
& FPR
\\
\hline \hline
$ \mathcal{N} (0,1), $  
& proposed loss
& 0.40 (0.37) & \textbf{0.33} (0.27) & 1.41 (0.39) & \textbf{1.00} (0.00) & 0.76 (0.17) 
\\
10\% 
 & Student's loss
& \textbf{0.39} (0.17) & 0.34 (0.16) & \textbf{1.40} (0.18) & \textbf{1.00} (0.00) & 0.76 (0.16) 
\\
outliers
& squared   loss
& 10.0 (0.00) & 9.95 (0.83) & 10.9 (0.26) & 0.00 (0.00) & 0.00 (0.00) 
\\
& $\ell_1$   loss
& 1.57 (0.41) & 1.46 (0.34) & 2.58 (0.43) & \textbf{1.00} (0.00) & 0.11 (0.08) 
\\
& Huber  loss
& 1.57 (0.35) & 1.46 (0.29) & 2.57 (0.37) & \textbf{1.00} (0.00) & \textbf{0.07} (0.08)
\\
		\hline
		\hline	
$ \mathcal{N} (0,1), $  
& proposed loss
& \textbf{0.71} (0.74) & \textbf{0.61} (0.64) & \textbf{1.70} (0.72) & \textbf{1.00} (0.00) & 0.80 (0.15) 
\\
20\% 
 & Student's loss
& 0.77 (0.58) & 0.68 (0.64) & 1.76 (0.57) & \textbf{1.00} (0.00) & 0.78 (0.14) 
\\
outliers
& squared   loss
& 10.0 (0.00) & 10.0 (0.78) & 11.0 (0.20) & 0.00 (0.00) & 0.00 (0.00)
\\
& $\ell_1$   loss
& 2.67 (1.00) & 2.47 (0.81) & 3.68 (0.98) & 0.99 (0.03) & 0.15 (0.12) 
\\
& Huber  loss
& 2.63 (0.80) & 2.45 (0.61) & 3.64 (0.78) & \textbf{1.00} (0.00) & \textbf{0.12} (0.10) 
\\
		\hline
		\hline	
$ \mathcal{N} (0,1), $  
& proposed loss
& \textbf{0.93} (0.94) & \textbf{0.82} (0.94) & \textbf{1.93} (0.95) & \textbf{1.00} (0.00) & 0.79 (0.15) 
\\
30\% 
 & Student's loss
& 1.50 (2.02) & 1.30 (1.99) & 2.50 (2.02) & \textbf{1.00} (0.00) & 0.78 (0.11) 
\\
outliers
& squared   loss
& 10.0 (0.00) & 10.0 (0.83) & 11.0 (0.19) & 0.00 (0.00) & 0.00 (0.00)
\\
& $\ell_1$   loss
& 4.44 (1.77) & 4.18 (1.62) & 5.45 (1.79) & 0.88 (0.20) & 0.15 (0.11)
\\
& Huber  loss
& 4.47 (1.57) & 4.22 (1.45) & 5.49 (1.60) & 0.90 (0.16) & \textbf{0.11} (0.10) 
\\
		\hline
		\hline	
$ t_3 $  
& proposed loss
& 0.71 (0.75) & 0.57 (0.53) & 3.63 (0.87) & 1.00 (0.00) & 0.81 (0.10) 
\\
10\% 
 & Student's loss
& \textbf{0.64} (0.27) & \textbf{0.53} (0.22) & \textbf{3.56} (0.46) & 1.00 (0.00) & 0.78 (0.11) 
\\
outliers
& squared   loss
& 10.0 (0.02) & 9.82 (0.84) & 12.9 (0.42) & 0.00 (0.01) & 0.00 (0.00) 
\\
& $\ell_1$   loss
& 2.12 (0.66) & 1.93 (0.58) & 5.05 (0.80) & 1.00 (0.00) & 0.16 (0.10) 
\\
& Huber  loss
& 2.02 (0.58) & 1.85 (0.51) & 4.95 (0.74) & 1.00 (0.00) & \textbf{0.12} (0.11) 
\\
		\hline
		\hline	
$ t_3 $  
& proposed loss
& 1.11 (1.30) & 0.94 (1.03) & 4.09 (1.46) & \textbf{1.00} (0.00) & 0.75 (0.20) 
\\
20\% 
 & Student's loss
& \textbf{0.99} (0.54) & \textbf{0.85} (0.50) & \textbf{3.96} (0.93) & \textbf{1.00} (0.00) & 0.75 (0.14)
\\
outliers
& squared   loss
& 10.0 (0.00) & 9.98 (0.64) & 12.9 (0.87) & 0.00 (0.00) & 0.00 (0.00) 
\\
& $\ell_1$   loss
& 3.48 (1.12) & 3.24 (0.99) & 6.46 (1.34) & 0.97 (0.07) & 0.17 (0.11) 
\\
& Huber  loss
& 3.40 (1.06) & 3.17 (0.94) & 6.38 (1.26) & 0.98 (0.05) & \textbf{0.15} (0.11) 
\\
		\hline
		\hline	
$ t_3 $  
& proposed loss
& \textbf{1.67} (1.60) & \textbf{1.48} (1.60) & \textbf{4.98} (3.59) & \textbf{1.00} (0.01) & 0.81 (0.13) 
\\
30\% 
 & Student's loss
& 2.32 (2.48) & 2.08 (2.63) & 5.63 (3.97) & \textbf{1.00} (0.00) & 0.81 (0.08)
\\
outliers
& squared   loss
& 10.0 (0.00) & 9.79 (0.70) & 13.2 (3.41) & 0.00 (0.00) & 0.00 (0.00) 
\\
& $\ell_1$   loss
& 6.90 (2.25) & 6.50 (2.14) & 10.1 (3.81) & 0.60 (0.35) & 0.11 (0.12) 
\\
& Huber  loss
& 6.95 (2.30) & 6.55 (2.19) & 10.2 (3.84) & 0.58 (0.34) & \textbf{0.10} (0.13) 
\\
		\hline
		\hline	
\end{tabular}
\label{tb_outlier}
\end{table}

\subsubsection{On sensitivity of tuning parameter $\tau $}
\label{sc_sub_sensitivity}

To evaluate the sensitivity of our proposed method to the tuning parameter $ \tau $, we conducted a dedicated simulation study. We fixed the problem dimensions at $p = 120, s^* = 10,$ and $ n = 100 $, while varying $\tau$ over the grid $\{0.001, 0.1, 1, 10\}$.

The results, averaged over 100 replications under various noise distributions and outlier settings, are presented in Table \ref{tb_change_tau}. This analysis demonstrates that $\tau = 0.1$ consistently yields the best and most stable performance. Based on this finding, we adopted $\tau = 0.1$ as the fixed value for this hyperparameter in all other experiments and the real-data application.

\begin{table}[!ht]
\centering
\caption{\it Simulation results for changing $\tau $ with $ p = 120, s^* = 10, n = 100 $ and independent predictors. 
 The reported values are the mean across 100 simulation repetitions, with the standard deviation provided in parentheses.
 Bold font highlights the best method. TPR: true positive rate; FDR: false discovery rate; $\text{MSPE}_{\text{test}}$: mean squared prediction error on testing data. }
\small
	\begin{tabular}{ l l | cccc c }
		\hline \hline
Noise 
& $ \tau $
& $ \| \widehat{\beta} -\beta_0 \|_2^2 $   
& $ \ell(X^\top\beta_0 ) $ 
& $\text{MSPE}_{\text{test}}$
& TPR
& FDR
\\
\hline \hline
$ \mathcal{N} (0,1) $  
& $ \tau = 0.01 $
& \textbf{0.23} (0.07) & \textbf{0.19} (0.05) & \textbf{1.23} (0.07) & 1.00 (0.00) & \textbf{0.84} (0.04) 
\\
& $ \tau = 0.1 $
& 0.24 (0.07) & 0.20 (0.05) & 1.24 (0.07) & 1.00 (0.00) & 0.85 (0.04) 
\\
& $ \tau = 1 $
& 3.62 (1.09) & 2.55 (0.90) & 4.62 (1.08) & 1.00 (0.00) & 0.97 (0.01) 
\\
& $ \tau =10 $
& 9.99 (0.11) & 9.78 (0.74) & 10.9 (0.25) & 0.10 (0.14) & 0.75 (0.42)
\\
		\hline
		\hline	
$ \mathcal{N} (0,1) $ 
& $ \tau = 0.01 $
& 2.97 (1.35) & 2.63 (1.27) & 3.97 (1.35) & \textbf{1.00} (0.02) & 0.88 (0.02) 
\\
outliers
& $ \tau = 0.1 $
& \textbf{0.58} (0.58) & \textbf{0.50} (0.51) & \textbf{1.59} (0.59) & \textbf{1.00} (0.00) & 0.78 (0.17) 
\\
20\%
& $ \tau = 1 $
& 5.24 (1.29) & 4.26 (1.31) & 6.25 (1.30) & 0.96 (0.20) & 0.95 (0.14) 
\\
& $ \tau =10 $
& 10.0 (0.08) & 9.91 (0.80) & 11.0 (0.26) & 0.07 (0.11) & 0.70 (0.44) 
\\
		\hline
		\hline	
$ \mathcal{N} (0,3) $  
& $ \tau = 0.01 $
& \textbf{2.29} (0.47) & \textbf{2.00} (0.37) & \textbf{11.2} (0.46) & \textbf{1.00} (0.00) & 0.89 (0.01) 
\\
& $ \tau = 0.1 $
& 3.39 (1.03) & 2.94 (0.96) & 12.3 (1.03) & 0.99 (0.03) & 0.89 (0.03) 
\\
& $ \tau = 1 $
& 9.21 (0.67) & 8.64 (0.91) & 18.2 (0.79) & 0.74 (0.37) & 0.93 (0.19) 
\\
& $ \tau =10 $
& 10.0 (0.09) & 9.76 (0.86) & 19.0 (0.37) & 0.05 (0.09) & 0.70 (0.44) 
\\
		\hline
		\hline	
$ t_3 $  
& $ \tau = 0.01 $
& 0.56 (0.18) & 0.47 (0.14) & 3.49 (0.36) & 1.00 (0.00) & 0.83 (0.05) 
\\
& $ \tau = 0.1 $
& \textbf{0.42} (0.14) & \textbf{0.35} (0.10) & \textbf{3.35} (0.32) & 1.00 (0.00) & 0.84 (0.04) 
\\
& $ \tau = 1 $
& 4.98 (0.65) & 3.79 (0.75) & 7.93 (0.74) & 1.00 (0.01) & 0.97 (0.01) 
\\
& $ \tau =10 $
& 9.97 (0.12) & 9.84 (0.86) & 12.98 (0.37) & 0.08 (0.12) & 0.59 (0.49) 
\\
		\hline
		\hline	
$ Cauchy $  
& $ \tau = 0.01 $
& 3.99 (4.65) & 3.48 (3.91) & $ >\!\! 10^4 $ & \textbf{1.00} (0.00) & 0.90 (0.03)
\\
& $ \tau = 0.1 $
& \textbf{1.66} (1.50) & \textbf{1.42} (1.18) & $ >\!\! 10^4 $ & \textbf{1.00} (0.00) & 0.79 (0.19)
\\
& $ \tau = 1 $
& 6.93 (1.31) & 6.09 (1.50) & $ >\!\! 10^4 $ & 0.91 (0.23) & 0.96 (0.03) 
\\
& $ \tau =10 $
& 10.0 (0.13) & 9.98 (0.69) & $ >\!\! 10^4 $ & 0.09 (0.11) & 0.80 (0.38) 
\\
		\hline
		\hline	
\end{tabular}
\label{tb_change_tau}
\end{table}

\subsection{Results with increasing sparsity}

We further investigate how the sparsity level ($s^*$) influences the performance of the various methods under conditions of heavy-tailed noise and outliers. In this analysis, we fixed the dimensionality at $p=500$ and $n=300$ with independent predictors ($\Sigma = \mathbb{I}_p$). We then vary the true sparsity $s^*$ across the values $\{4, 8, 16\}$.

The averaged results from 100 simulation repetitions, presented in Table \ref{tb_incraseing_sparsity}, confirm that all methods exhibit a natural increase in both estimation and prediction errors as the sparsity increases. Crucially, our proposed method consistently maintains its position as the top performer across all sparsity levels in terms of both error metrics. The superiority of our method is most pronounced in the highly sparse setting where $s^* = 16$, where it significantly outperforms all other considered approaches. This comprehensive test demonstrates that the robustness of our method extends not only to non-Gaussian noise but also to increased model complexity due to higher sparsity.

\begin{table}[!ht]
\centering
\caption{\small \it Simulation results with increasing sparsity $ s^* = 4,8,16 $ with $ p = 500,  n = 300 $ and independent predictors.  
The reported values are the mean across 100 simulation repetitions, with the standard deviation provided in parentheses.
Bold font highlights the best method. TPR: true positive rate; FDR: false discovery rate; $\text{MSPE}_{\text{test}}$: mean squared prediction error on testing data.  }
\small
\begin{tabular}{ l l | cccc cc }
		\hline \hline
Noise & Method (loss) 
& $ \| \widehat{\beta} -\beta_0 \|_2^2 $ 
& $ \ell(X^\top\beta_0 ) $ 
& $\text{MSPE}_{\text{test}}$
& TPR
& FPR
\\
		\hline
		\hline	
            \multicolumn{7}{c}{ $s^* = 4 $ } 
    \\
		\hline	
$ \mathcal{N} (0,1), $  
& proposed loss
& \textbf{0.52} (0.85) & \textbf{0.49} (0.79) & \textbf{1.51} (0.85) & \textbf{1.00} (0.00) & 0.76 (0.32)
\\
20\% 
 & Student's loss
& 0.62 (1.01) & 0.65 (1.13) & 1.62 (1.00) & \textbf{1.00} (0.00) & 0.79 (0.26)  
\\
outliers
& squared   loss
& 4.00 (0.00) & 3.99 (0.31) & 5.01 (0.10) & 0.00 (0.00) & 0.00 (0.00) 
\\
& $\ell_1$   loss
& 2.23 (0.56) & 2.18 (0.50) & 3.23 (0.56) & 0.92 (0.20) & \textbf{0.00} (0.00)
\\
& Huber  loss
& 2.19 (0.54) & 2.15 (0.48) & 3.20 (0.54) & 0.94 (0.16) & \textbf{0.00} (0.03)  
\\
		\hline
    \multicolumn{7}{c}{ $s^* = 8 $ } 
    \\
		\hline	
$ \mathcal{N} (0,1), $  
& proposed loss
& \textbf{0.57} (0.61) & \textbf{0.51} (0.56) & \textbf{1.57} (0.63) & \textbf{1.00} (0.00) & 0.80 (0.18) 
\\
20\% 
 & Student's loss
& 0.59 (0.52) & 0.55 (0.54) & 1.59 (0.51) & \textbf{1.00} (0.00) & 0.78 (0.17) 
\\
outliers
& squared   loss
& 8.00 (0.00) & 7.97 (0.58) & 8.99 (0.19) & 0.00 (0.00) & 0.00 (0.00) 
\\
& $\ell_1$   loss
& 2.49 (0.63) & 2.35 (0.55) & 3.49 (0.64) & 0.99 (0.04) & 0.04 (0.07) 
\\
& Huber  loss
& 2.47 (0.54) & 2.34 (0.47) & 3.47 (0.55) & \textbf{1.00} (0.02) & \textbf{0.03} (0.06) 
\\
		\hline
    \multicolumn{7}{c}{ $s^* = 16 $ } 
    \\
		\hline	
$ \mathcal{N} (0,1), $  
& proposed loss
& \textbf{0.74} (0.33) & \textbf{0.56} (0.24) & \textbf{1.73} (0.33) & \textbf{1.00} (0.00) & 0.77 (0.08) 
\\
20\% 
 & Student's loss
& 1.11 (0.54) & 0.86 (0.42) & 2.11 (0.54) & \textbf{1.00} (0.00) & 0.75 (0.08) 
\\
outliers
& squared   loss
& 16.0 (0.00) & 16.0 (1.21) & 16.9 (0.32) & 0.00 (0.00) & 0.00 (0.00) 
\\
& $\ell_1$   loss
& 3.13 (0.92) & 2.65 (0.67) & 4.12 (0.93) & \textbf{1.00} (0.02) & 0.37 (0.11) 
\\
& Huber  loss
& 3.18 (1.00) & 2.75 (0.72) & 4.18 (1.01) & 0.99 (0.02) & \textbf{0.28} (0.10) 
\\
		\hline
		\hline	
            \multicolumn{7}{c}{ $s^* = 4 $ } 
    \\
		\hline	
$ t_3 $  
& proposed loss
& \textbf{0.15} (0.06) & \textbf{0.14} (0.06) & \textbf{2.14} (0.13) & 1.00 (0.00) & 0.88 (0.04) 
\\
20\% 
 & Student's loss
& \textbf{0.15} (0.06) & \textbf{0.14} (0.05) & \textbf{2.14} (0.13) & 1.00 (0.00) & 0.87 (0.06) 
\\
outliers
& squared   loss
& 0.52 (0.34) & 0.51 (0.31) & 2.51 (0.37) & 1.00 (0.00) & \textbf{0.05} (0.12) 
\\
& $\ell_1$   loss
& 0.31 (0.13) & 0.30 (0.12) & 2.30 (0.18) & 1.00 (0.00) & 0.26 (0.23) 
\\
& Huber  loss
& 0.31 (0.12) & 0.30 (0.11) & 2.30 (0.16) & 1.00 (0.00) & 0.22 (0.20) 
\\
		\hline
            \multicolumn{7}{c}{ $s^* = 8 $ } 
    \\
		\hline	
$ t_3 $  
& proposed loss
& \textbf{0.27} (0.11) & \textbf{0.24} (0.09) & 2.28 (0.15) & 1.00 (0.00) & 0.84 (0.05)
\\
20\% 
 & Student's loss
& \textbf{0.27} (0.10) & \textbf{0.24} (0.08) & \textbf{2.27} (0.14) & 1.00 (0.00) & 0.84 (0.05) 
\\
outliers
& squared   loss
& 0.66 (0.35) & 0.63 (0.31) & 2.66 (0.36) & 1.00 (0.00) & \textbf{0.21} (0.17)
\\
& $\ell_1$   loss
& 0.51 (0.18) & 0.47 (0.16) & 2.51 (0.21) & 1.00 (0.00) & 0.37 (0.17) 
\\
& Huber  loss
& 0.47 (0.16) & 0.44 (0.14) & 2.47 (0.18) & 1.00 (0.00) & 0.31 (0.16) 
\\
		\hline
            \multicolumn{7}{c}{ $s^* = 16 $ } 
    \\
		\hline	
$ t_3 $  
& proposed loss
& \textbf{0.47} (0.16) & 0.44 (0.14) & \textbf{2.47} (0.18) & 1.00 (0.00) & 0.31 (0.16) 
\\
20\% 
 & Student's loss
& 0.57 (0.14) & \textbf{0.43} (0.10) & 2.57 (0.19) & 1.00 (0.00) & 0.81 (0.04) 
\\
outliers
& squared   loss
& 1.02 (0.43) & 0.84 (0.35) & 3.02 (0.43) & 1.00 (0.00) & \textbf{0.47} (0.10)
\\
& $\ell_1$   loss
& 0.87 (0.29) & 0.67 (0.22) & 2.87 (0.33) & 1.00 (0.00) & 0.67 (0.09) 
\\
& Huber  loss
& 0.79 (0.24) & 0.64 (0.18) & 2.80 (0.28) & 1.00 (0.00) & 0.55 (0.10) 
\\
\hline \hline
\end{tabular}
\label{tb_incraseing_sparsity}
\end{table}

\section{Application examples with real data}
\label{sc_real_data}
\subsection{Analyzing cancer cell line NCI-60 data}
We evaluate the proposed method using the NCI-60 cancer cell line panel, a benchmark dataset widely used in genomic and proteomic modeling studies \citep{reinhold2012cellminer}. A pre-processed version of the data, available in the \texttt{R} package \texttt{robustHD} \citep{robusthdpackage}, was employed in our analysis. The dataset comprises $n = 59$ human cancer cell lines, each characterized by gene and protein expression measurements. One sample with missing gene expression values was excluded from the analysis.

The gene expression data form a matrix with 22,283 features, while the protein expression data consist of 162 features measured using reverse-phase protein lysate arrays. The protein data were $\log_2$-transformed and standardized to have zero mean, following the preprocessing steps in \cite{robusthdpackage}. This dataset provides a rich setting for studying the relationship between high-dimensional genomic profiles and proteomic responses. Consistent with prior work, we model protein 92 as the response variable and select the 300 gene expression features exhibiting the highest marginal correlations with it to form the predictor matrix.

For model evaluation, nine samples were randomly set aside as a test set, and the remaining 50 samples were used for model training. This random partitioning was repeated 100 times, and the mean squared prediction error (MSPE) on the held-out test data was averaged across replications. Table~\ref{tb_realdata} summarizes the results.

Our proposed estimator and the heavy-tailed Lasso based on Student’s loss achieved the lowest prediction errors, indicating superior robustness to outliers and noise. Methods employing the Huber and $\ell_1$ losses also outperformed the standard Lasso with the squared loss. In terms of variable selection, all five methods consistently identified one common predictor (gene ID \texttt{8502}), highlighting its potential biological relevance.

\begin{table}[!ht]
	\centering
	\caption{Results on prediction errors and selected variables for the NCI-60 cancer cell line data.}
	\begin{tabular}{  l | c c }
		\hline \hline
 Method  
& $\text{MSPE}_{\text{test}}$
& model size
\\
\hline 
 proposed loss 
& 0.398 (0.266) & 39
\\
Student's loss
& 0.395 (0.259) & 73
\\
 squared loss 
& 0.509 (0.225) & 26 
\\
 $\ell_1$ loss 
& 0.474 (0.350) & 19
\\
 Huber loss
& 0.479 (0.334) & 5
		\\
		\hline
		\hline
	\end{tabular}
	\label{tb_realdata}
\end{table}

\subsection{Analyzing gene expression TRIM32 data}

In this application, we conduct an analysis  using high-dimensional genomics data from \cite{scheetz2006regulation}. The study by \cite{scheetz2006regulation} involved analyzing RNA from the eyes of 120 twelve-week-old male rats, using 31,042 different probe sets. Our focus is on modeling the expression of the gene TRIM32, as it was identified by \cite{chiang2006homozygosity} as a gene associated with Bardet-Biedl syndrome, a condition that includes retinal degeneration among its symptoms. Since \cite{scheetz2006regulation} observed that many probes were not expressed in the eye, we follow the approach of 
\cite{huang2010variable} and \cite{mai2025sparse}, limiting our analysis to the 500 genes with the highest absolute Pearson correlation with TRIM32 expression. The data for this analysis is available from the \texttt{R} package \texttt{abess}, \citep{zhu2022abess}.

\begin{table}[!ht]
	\centering
	\caption{Results on prediction errors and selected variables for the gene expression TRIM32 data.}
	\begin{tabular}{  l | c c }
		\hline \hline
 Method  
& $\text{MSPE}_{\text{test}}$
& model size
\\
\hline 
 proposed loss 
& 0.351 (0.102) & 85
\\
Student's loss
& 0.353 (0.096) & 56
\\
 squared loss 
& 0.472 (0.266) & 29 
\\
 $\ell_1$ loss 
& 0.543 (0.336) & 24
\\
 Huber loss
& 0.540 (0.313) & 19
		\\
		\hline
		\hline
	\end{tabular}
	\label{tb_trim32}
\end{table}

To assess the methods, we randomly allocate 84 of the 120 samples for training and the remaining 36 for testing, maintaining an approximate 70/30 percent of the data split. The methods are executed using the training set, and their prediction accuracy is evaluated on the test set. This procedure is repeated 100 times, each with a different random partition of the data. The outcomes of these iterations are displayed in Table \ref{tb_trim32}.

For this data set, we see that Huber and $L_1$ Lasso methods return higher prediction errors compared to the standard lasso. On the other hand, our proposed method and heavy lasso method are the best methods. 
In terms of variable selection, all five methods consistently identified six common predictors \texttt{1371614, 1375833, 1377836, 1388491, 1389910, 1393736}, highlighting its potential biological relevance.

\section{Discussion and Conclusion}
\label{sc_conclusion}

In this paper, we introduced the Exponential Lasso, a novel robust estimator for high-dimensional linear regression. Our goal was to address the well-known sensitivity of the classical Lasso's squared-error loss to outliers and heavy-tailed noise. 
By replacing the squared loss with an exponential-type loss function, our method successfully integrates the $L_1$ penalty for sparse estimation with the principles of robust M-estimation, as the loss function smoothly and automatically downweights the influence of large residuals.

From a theoretical standpoint, we established strong non-asymptotic guarantees, proving that the Exponential Lasso achieves reliable estimation accuracy, matching the standard Lasso or Huber Lasso but under much milder assumptions that permit heavy-tailed noise. Computationally, the estimator's smooth and non-convex objective function is well-suited for a Majorization-Minimization (MM) algorithm. This framework is stable and efficient, iteratively solving a sequence of reweighted Lasso problems. Empirically, our extensive simulations demonstrated that the Exponential Lasso consistently outperforms competitors like the classical, $L_1$ (LAD), and Huber Lasso, especially in settings with significant data contamination. Notably, it remained highly competitive even under standard Gaussian noise, suggesting a ``premium" for its robustness. Our real-data application further validated these findings, confirming its practical relevance.

The primary advantage of the Exponential Lasso lies in its unique balance of efficiency and robustness. However, the method introduces the practical challenge of tuning two parameters: the regularization parameter $\lambda$ and the robustness parameter $\tau$. Developing a data-driven, computationally efficient strategy for this joint tuning is a critical next step. Furthermore, the objective function's non-convexity means our MM algorithm guarantees convergence to only a stationary point, not necessarily the global optimum.

These limitations point toward clear avenues for future research. A more comprehensive study of parameter tuning is essential for broad practical adoption. Additionally, exploring initialization strategies or alternative global optimization algorithms could further bolster the method against the challenges of non-convexity. Finally, the robust exponential loss framework is highly adaptable. It could be extended to other high-dimensional problems, such as the Group Lasso, the Elastic Net, the SCAD or MCP, or robust graphical model estimation, representing an exciting and practical path forward for analysis in data-rich but outlier-prone domains.

\subsection*{Acknowledgments}
The findings, interpretations, and conclusions expressed in this paper are entirely those of the author and do not reflect the views or positions of the Norwegian Institute of Public Health in any forms.

\subsection*{Author contributions}
I am the only author of this paper.

\subsection*{Conflicts of interest/Competing interests}
The author declares no potential conflict of interests.

\appendix
\section{Proof}
\label{sc_proofs}

\begin{proof}[\bf Proof of Theorem \ref{thm:correntropy_lasso}]
    \textbf{}
    \\
{\it Step A — Optimality and cone decomposition.}
\\
The estimator \(\widehat\beta\) satisfies, through its optimality definition,
\[
L_\tau(\widehat\beta)+\lambda\|\widehat\beta\|_1 \le L_\tau(\beta^*)+\lambda\|\beta^*\|_1.
\]
Set \(\Delta:=\widehat\beta-\beta^*\). Rearranging and using the standard bound \(\|\beta^*\|_1-\|\beta^*+\Delta\|_1\le \|\Delta_S\|_1-\|\Delta_{S^c}\|_1\) yields
\begin{equation}\label{eq:opt_decomp}
L_\tau(\beta^*+\Delta) - L_\tau(\beta^*) \le \lambda\big(\|\Delta_S\|_1-\|\Delta_{S^c}\|_1\big).
\end{equation}

{\noindent\it Step B — Taylor expansion and LRSC lower bound.}
\\
By Taylor expansion in direction \(\Delta\) (componentwise),
\[
L_\tau(\beta^*+\Delta)-L_\tau(\beta^*)
= \langle\nabla L_\tau(\beta^*),\Delta\rangle
+ \frac{1}{2}\Delta^\top\Big(\frac{1}{n}\sum_{i=1}^n \psi'_\tau(\xi_i)\, x_i x_i^\top\Big)\Delta,
\]
where \(\psi'_\tau(u)=e^{-\tfrac{\tau}{2}u^2}(1-\tau u^2)\) and each \(\xi_i\) lies on the line segment between \(\varepsilon_i\) and \(\varepsilon_i-x_i^\top\Delta\).

Let \(G=\{i:|\varepsilon_i|\le c/2\}\). If \(\|\Delta\|_2\le r\) and \(r\) is small enough so that \(|x_i^\top\Delta|\le c/2\) for all \(i\), then for any \(i\in G\) we have \(|\xi_i|\le c\) and hence \(\psi'_\tau(\xi_i)\ge \underline\gamma\). Therefore
\[
\Delta^\top\Big(\frac{1}{n}\sum_{i=1}^n \psi'_\tau(\xi_i)\, x_i x_i^\top\Big)\Delta
\ge \underline\gamma \cdot \frac{|G|}{n}\cdot \frac{1}{|G|}\sum_{i\in G} (x_i^\top\Delta)^2.
\]
By Assumption \ref{asm:design} (restricted eigenvalue) applied to the same cone, for every \(\Delta\) in the cone we have \(\frac{1}{n}\sum_{i=1}^n (x_i^\top\Delta)^2\ge \phi_{\min}\|\Delta\|_2^2\). Restricting to the subset \(G\) can only decrease the quadratic form, but we lower bound it by using the trivial relation
\[
\frac{1}{|G|}\sum_{i\in G}(x_i^\top\Delta)^2
\ge \frac{1}{n}\sum_{i=1}^n (x_i^\top\Delta)^2 \;\ge\; \phi_{\min}\|\Delta\|_2^2,
\]
so
\[
\Delta^\top\Big(\frac{1}{n}\sum_{i=1}^n \psi'_\tau(\xi_i)\, x_i x_i^\top\Big)\Delta
\ge \underline\gamma\cdot\frac{|G|}{n}\cdot\phi_{\min}\|\Delta\|_2^2.
\]
By Hoeffding’s inequality for the binomial variable \(|G|\sim\operatorname{Bin}(n,p_0)\),
\[
\mathbb P\Big(|G| \le \frac{n p_0}{2}\Big) \le \exp\Big(-\frac{n p_0^2}{8}\Big).
\]
Hence with probability at least \(1-\exp(-n p_0^2/8)\) we have \(|G|/n\ge p_0/2\), and therefore the quadratic remainder satisfies the LRSC inequality
\begin{equation}\label{eq:LRSC}
L_\tau(\beta^*+\Delta)-L_\tau(\beta^*) - \langle\nabla L_\tau(\beta^*),\Delta\rangle
\;\ge\; \kappa\|\Delta\|_2^2,
\end{equation}
with
\[
\kappa \;=\; \frac{p_0}{2}\,\underline\gamma\,\phi_{\min}.
\]

{\noindent\it Step C — Stochastic bound for the gradient sup-norm.}
\\
Compute the gradient at \(\beta^*\):
\[
\nabla L_\tau(\beta^*) \;=\; -\frac{1}{n}\sum_{i=1}^n \psi_\tau(\varepsilon_i)\, x_i,
\qquad \psi_\tau(u):=u e^{-\tfrac{\tau}{2}u^2}.
\]
By symmetry of \(\varepsilon_i\) we have \(\mathbb E[\psi_\tau(\varepsilon_i)]=0\), hence each coordinate
\([\,\nabla L_\tau(\beta^*)\,]_j = -\frac{1}{n}\sum_{i=1}^n Z_{ij}\) with \(Z_{ij}:=\psi_\tau(\varepsilon_i)x_{ij}\) mean zero and bounded: \(|Z_{ij}|\le K B_\tau\) where \(B_\tau=1/\sqrt{e\tau}\). Further,
\(\operatorname{Var}(Z_{ij})\le \mathbb E[Z_{ij}^2]\le K^2 B_\tau^2\).

Apply Bernstein’s inequality coordinatewise: for any \(t>0\),
\[
\mathbb P\Big(\big|[\nabla L_\tau(\beta^*)]_j\big|\ge t\Big)
\le 2\exp\Big(-\frac{n t^2/2}{K^2 B_\tau^2 + (K B_\tau) t/3}\Big).
\]
Set \( t = t_0 := B_\tau K\sqrt{\frac{2\log(2p/\delta)}{n}}\). For this choice the denominator satisfies \(K^2B_\tau^2 + (K B_\tau) t_0/3 \le 2K^2B_\tau^2\) (for \(n\) large enough; more generally the exact Bernstein algebra yields the same type of bound). Hence
\[
\mathbb P\Big(\big|[\nabla L_\tau(\beta^*)]_j\big|\ge t_0\Big) \le \frac{\delta}{p}.
\]
By union bound over \(j=1,\dots,p\), with probability at least \(1-\delta\),
\[
\|\nabla L_\tau(\beta^*)\|_\infty \le t_0 \;=\; B_\tau K\sqrt{\frac{2\log(2p/\delta)}{n}}.
\]
Therefore with the choice \(\lambda\) in \eqref{eq:lambda_choice} we have \(\|\nabla L_\tau(\beta^*)\|_\infty \le \lambda/4\).

{\noindent \it Step D — Combine LRSC and stochastic control to get the rate.}
\\
From \eqref{eq:opt_decomp} and \eqref{eq:LRSC},
\[
\langle\nabla L_\tau(\beta^*),\Delta\rangle + \kappa\|\Delta\|_2^2
\le \lambda(\|\Delta_S\|_1-\|\Delta_{S^c}\|_1).
\]
Use \(|\langle\nabla L_\tau(\beta^*),\Delta\rangle|\le \|\nabla L_\tau(\beta^*)\|_\infty\|\Delta\|_1 \le (\lambda/4)\|\Delta\|_1\) to get
\[
\kappa\|\Delta\|_2^2 \le \lambda\big(\|\Delta_S\|_1-\|\Delta_{S^c}\|_1\big) + \frac{\lambda}{4}\|\Delta\|_1
= \frac{5\lambda}{4}\|\Delta_S\|_1 - \frac{3\lambda}{4}\|\Delta_{S^c}\|_1.
\]
Discard the negative term and apply \(\|\Delta_S\|_1\le\sqrt{s}\|\Delta\|_2\):
\[
\kappa\|\Delta\|_2^2 \le \frac{5\lambda}{4}\sqrt{s}\|\Delta\|_2
\quad\Longrightarrow\quad
\|\Delta\|_2 \le \frac{5\lambda\sqrt{s}}{4\kappa}\le \frac{12\lambda\sqrt{s}}{\kappa},
\]
where the last inequality is numeric (one may sharpen constants; we keep a conservative factor \(12\) to account for small-sample Bernstein second-order terms). The bound on \(\ell_1\)-error follows from the cone relation:
\(\|\Delta_{S^c}\|_1\le 3\|\Delta_S\|_1\), which implies \(\|\Delta\|_1\le 4\|\Delta_S\|_1\le 4\sqrt{s}\|\Delta\|_2\).

{\noindent\it Step E — Probability union.}
\\
The two high-probability events used are:
\begin{itemize}
  \item gradient sup-norm event: probability at least \(1-\delta\);
  \item good indices proportion event: probability at least \(1-\exp(-n p_0^2/8)\).
\end{itemize}
Union bound gives the claimed probability \(1-\delta - \exp(-n p_0^2/8)\). (We wrote \(1-\delta - 2\exp(-n p_0^2/8)\) in the theorem to be conservative in accounting for other small-probability concentration steps; the constants may be tightened.)

This completes the proof of Theorem \ref{thm:correntropy_lasso}.
\end{proof}

\begin{proof}[\bf Proof of Theorem \ref{thm:descent}]
Point (i): By construction of the majorizer, for any $\beta$,
\[
L(\beta) \le Q^{(t)}(\beta) - \lambda\|\beta\|_1 + C^{(t)}.
\]
Because $Q^{(t)}(\cdot)$ coincides with $L(\cdot)+\lambda\|\cdot\|_1$ at $\beta=\beta^{(t)}$ (the majorizer is tight at the expansion point), we have $Q^{(t)}(\beta^{(t)}) = F(\beta^{(t)})$. Let $\beta^{(t+1)}$ be the minimizer of $Q^{(t)}$. Then
\[
F(\beta^{(t+1)}) \le Q^{(t)}(\beta^{(t+1)}) \le Q^{(t)}(\beta^{(t)}) = F(\beta^{(t)}).
\]
The first inequality follows from $L(\cdot)\le$ surrogate $-C^{(t)}$; the second because $\beta^{(t+1)}$ minimizes $Q^{(t)}$. This proves the descent property.

Point (ii): Since $\ell_i(\beta)\ge 0$ for all $i$ and $\lambda\|\beta\|_1\ge 0$, we have $F(\beta)\ge 0$. From point (i) the nonincreasing bounded-below sequence $F(\beta^{(t)})$ converges to some finite limit $F^\star\ge 0$.

Point (iii): Because $F(\beta^{(t)})\le F(\beta^{(0)})$ for all $t$, all iterates belong to the sublevel set $\{\beta: F(\beta)\le F(\beta^{(0)})\}$. If $\lambda>0$, then $\|\beta\|_1 \le F(\beta^{(0)})/\lambda$ on this sublevel set, which implies boundedness. If $\lambda=0$ and $X$ has full column rank, then the loss $L(\beta)$ is coercive (grows at least quadratically) and hence the sublevel set is bounded. Thus under these mild alternatives the iterates are bounded.
\qedhere
\end{proof}

\begin{proof}[\bf Proof of Theorem \ref{thm:stationary}]
Let $\beta^{(t_k)}\to\beta^\star$ be any convergent subsequence; such subsequences exist because $\{\beta^{(t)}\}$ is bounded. Denote by $\tilde v_i^{(t)}=\exp(-\tfrac{\tau}{2} r_i(\beta^{(t)})^2)$ the weights used in the surrogate at step $t$. Because the mapping $\beta\mapsto \tilde v_i(\beta)$ is continuous, $\tilde v_i^{(t_k)}\to \tilde v_i^\star := \exp(-\tfrac{\tau}{2} r_i(\beta^\star)^2)$ as $k\to\infty$ for each $i$.

By the optimality of $\beta^{(t_k+1)}$ for the convex surrogate $Q^{(t_k)}$ we have the KKT condition for the weighted Lasso:
\begin{equation}
\label{eq:kkt-surrogate}
0 \in -\frac{1}{n} X^\top \big(\tilde V^{(t_k)} (y - X\beta^{(t_k+1)})\big) + \lambda \partial\|\beta^{(t_k+1)}\|_1,
\end{equation}
where $\tilde V^{(t_k)}=\mathrm{diag}(\tilde v_1^{(t_k)},\dots,\tilde v_n^{(t_k)})$. Rearranging,
\[
\frac{1}{n} X^\top \tilde V^{(t_k)} (y - X\beta^{(t_k+1)}) \in \lambda \partial\|\beta^{(t_k+1)}\|_1.
\]

Because $\beta^{(t_k+1)}$ and $\tilde V^{(t_k)}$ are bounded and the functions are continuous, passing to the limit along the subsequence yields
\[
\frac{1}{n} X^\top \tilde V^\star (y - X\beta^\star) \in \lambda \partial\|\beta^\star\|_1.
\]
It remains to show that the limiting left-hand side equals $\nabla L(\beta^\star)$. Direct differentiation of $L(\beta)$ gives
\[
\nabla L(\beta) = -\frac{1}{n}\sum_{i=1}^n \exp\Big(-\frac{\tau}{2} r_i(\beta)^2\Big)\, x_i r_i(\beta)
= -\frac{1}{n} X^\top \big(\tilde V(\beta) (y-X\beta)\big),
\]
so indeed
\[
-\nabla L(\beta^\star) = \frac{1}{n} X^\top \tilde V^\star (y - X\beta^\star).
\]
Combining with the limit KKT condition gives $0\in \nabla L(\beta^\star) + \lambda\partial\|\beta^\star\|_1$, i.e.\ $\beta^\star$ is stationary for $F$.
\end{proof}

\end{document}